\documentclass[twoside,11pt]{article}
\usepackage{jmlr2e}
\usepackage{paralist}
\usepackage{times}
\usepackage{amsmath,amsfonts,amscd,amsthm,amssymb,eufrak,bbm,mathrsfs}
\usepackage{graphicx,caption,subcaption,epstopdf,float} 
\usepackage{epsfig}
\usepackage[square,comma,numbers]{natbib}
\usepackage{algorithm}
\usepackage{algpseudocode}
\algnotext{EndFor}
\usepackage{hyperref}
\usepackage[symbol]{footmisc}
\usepackage{array,booktabs,paralist}
\newtheorem{theorem}{Theorem}
\newtheorem{lemma}{Lemma}
\newtheorem{assumption}{Assumption}
\newtheorem{definition}{Definition}

\newtheorem{proposition}{Proposition}

\begin{document}
\title{Consistent Collective Matrix Completion under Joint Low Rank Structure}
\author{\name Suriya Gunasekar \email suriya@utexas.edu \AND \name Makoto Yamada
\email makotoy@yahoo-inc.com \AND \name Dawei Yin \email daweiy@yahoo-inc.com \AND Yi Chang \email yichang@yahoo-inc.com}

\maketitle

\begin{abstract}
We address the collective matrix completion problem of jointly recovering a  collection of matrices with shared structure from partial (and potentially noisy) observations. To ensure well--posedness of the problem, we impose a joint low rank structure, wherein each component matrix is low rank and the latent space of the low rank factors corresponding to each entity is shared across the entire collection. 
We first develop a rigorous algebra for representing and manipulating collective--matrix structure, and identify sufficient conditions for consistent estimation of collective matrices. We then propose a tractable convex estimator for solving the collective matrix completion problem, and provide the first non--trivial theoretical guarantees  for consistency of collective matrix completion. We show that under reasonable assumptions stated in Sec.~\ref{sec:Ass}, with high probability, the proposed estimator exactly recovers the true matrices whenever sample complexity requirements 
dictated by Theorem~\ref{thm:main} are met. The sample complexity requirement derived in the paper are optimum up to logarithmic factors, and significantly improve upon the requirements obtained by trivial extensions of standard matrix completion. Finally, we propose a scalable approximate algorithm to solve the proposed convex program, and  corroborate our results through  simulated experiments.
\end{abstract}
\section{Introduction}\label{sec:intro} 
Affinity relationships between a pair of \textit{entity types} (e.g.~users, movies, documents, explicit features, etc.) are often represented in a matrix form.
The standard matrix completion task of predicting the missing entries of a matrix from partial (and potentially noisy) observations is at the core of a wide range of applications including recommendation systems, recovering gene--protein interactions, and modeling text document collections, among others~\cite{koren2009matrix,dueck2005multi,xu2003document}. 
In many practical applications, data from multiple matrices often share correlated information, and leveraging the shared structure can potentially enhance performance. For example, in e--commerce applications, user preferences in multiple domains such as  news, ads, etc., and explicit user/item feature information such as demographics, social network, text description, etc., are made available in the form of a ``collection of matrices" sharing interactions among a common set of users/items. 

\textit{Collective matrix completion} involves  simultaneously completing one or more partially observed matrices by  leveraging  data from a set of correlated matrices. Each component matrix, also called a \textit{view}, represents pairwise affinity relation among $K$ types of \textit{entities}. We assume a \textit{joint low rank} structure, wherein each entity type $k$ has  a low dimensional latent factor representation $U_k$; and each view $v$ representing the affinity between entity types $k_1$ and $k_2$ is a low rank matrix given by $U_{k_1}U_{k_2}^\top$.  Leveraging such shared structure is especially attractive in scenarios where standard matrix completion typically fails, such as: 
\begin{inparaenum}[(i)]
\item \emph{Insufficient Data}: Data sparsity in one view can often be mitigated by augmenting data from related views. For example, in a multiple recommendation systems, user's interests can be better captured by combining data from multiple sources;
\item \emph{Cold Start}: Recommendation for new users/items with no prior ratings can be partially addressed in collective matrix completion using additional data like explicit user/item features. 
\end{inparaenum}

However, the problem of collective--matrix completion, like standard matrix completion, is statistically ill--posed as: (a) only a decaying fraction of the number of entries in a matrix are observed; (b) the observations are localized (e.g. individual matrix entries as opposed to random linear measurements). 
Recent works on matrix completion leverage the developments in high dimensional estimation~\cite{negahban2009unified,chandrasekaran2012convex,vershynin2014estimation, candesmathematics}, and propose statistically consistent tractable estimators under  low rank and other structural assumptions~\cite{candes2009exact,candes2010matrix,keshavan2010matrix,keshavan2010noise,recht2011simpler,gross2011recovering,negahban2012restricted,davenport2012bit,jain2013low,
 gunasekar2014exponential,chen2014coherent}.  However, to the best of our knowledge, optimal  sample complexity requirements for statistically consistent recovery of collective--matrices has not been previously analyzed.  
 
In this paper, we propose a convex estimator for collective matrix completion and provide the first non--trivial theoretical guarantees for consistent recovery of collective--matrices. In a close related work, Bouchard~et~al.~\cite{bouchard2013convex} propose the first convex estimator for collective matrix completion without analyzing the consistency of the estimate. 
In comparison to the analysis for standard matrix completion, several new challenges are encountered in collective matrix completion: \begin{asparaenum}[(a)] \item Trivial extensions of sample complexity from existing results on standard matrix completion are suboptimal as they do not consider the shared structure. Thus, fully leveraging the joint low--rank structure in the analysis is the key to obtain optimal sample complexity. \item Unlike matrices, for collective matrices with joint low rank structure, the entity factors $U_k$ are not always unique (upto signs and normalization). However, we observe that, under the assumptions in Sec.~\ref{sec:Ass}, even when $U_k$ are not unique, the $V$ relevant interactions are uniquely captured. \item For general collective--matrix structures, a joint factorization may not always exist (even with full rank), and further the proposed convex estimator can be badly behaved, we enforce Assumption~\ref{ass:g} to avoid these cases; although this assumption can  potentially be relaxed.\end{asparaenum}

To summarize our contributions:
\begin{asparaenum}[(i)]
\item In Sec.~\ref{sec:cmf} and \ref{sec:convexEstimator}, we develop a rigorous algebra for representing and manipulating collective--matrices. We identify sufficient conditions (Assumptions \ref{ass:lr}--\ref{ass:g}) under which consistent recovery is feasible, and propose a tractable convex estimator for collective matrix completion. 
\item We provide the first theoretical guarantee for consistent collective matrix completion (Theorem~\ref{thm:main}). Specifically, we show that for a subset of collective--matrix structures, with high probability, the proposed estimator exactly recovers the true matrices whenever the sample complexity satisfies  $\forall k$, $|\Omega_k|\sim O(n_kR\overline{\text{log}}{N})$, where $n_k$ is the number of entities of type $k$, $R$ is the joint rank of the collective matrices, and $|\Omega_k|$ is the expected number of observations corresponding to entity $k$. We note that these rates are optimal upto logarithmic factors.
\item 
Finally, while the proposed convex program can be solved by adapting the Singular Value Thresholding for Collective Matrix Completion (SVT--CMC) algorithm proposed by  Bouchard~et~al. \cite{bouchard2013convex,cai2010singular,toh2010accelerated}, this algorithm is not scalable to large datasets. As a minor contribution, we adapt Hazan's algorithm \cite{hazan2008sparse} to provide an approximate solution for the proposed convex program (Sec.~\ref{sec:algo}). The proposed algorithm has a significantly better per iteration complexity as compared to  SVT--CMC, and can be used to tradeoff accuracy for computation in large datasets. We conclude the paper by corroborating our results through experiments on simulated and real life datasets (Sec.~\ref{sec:exp}).
\end{asparaenum}

Besides the convex estimator, related work for collective matrix completion includes various non--convex estimators and probabilistic models. A seminal paper on low rank collective matrix factorization is the work by Singh~et~al.~\cite{singh2008relational}, wherein the views are parameterized by the shared latent factor representation. The latent factors are learnt by minimizing a regularized loss function over the estimates. 
 A Bayesian model for collective matrix factorization was also proposed by the same authors~\cite{singh2009efficient,singh2012bayesian}. 
  Collective matrix factorization is also related to applications involving multi--task learning and tensor factorization~\cite{long2006spectral,lippert2008relation,agarwal2011localized,yilmaz2011generalised,zhang2012multi}. 
For the special case of low rank matrix completion, besides the theoretical guarantees, there are plenty of equally significant work that propose effective and scalable algorithms, including max--margin matrix factorization \cite{srebro2004maximum}, alternating minimization \cite{koren2009matrix,zhou2008als}, and probabilistic models \cite{mnih2007probabilistic,salakhutdinov2008bayesian}, among others. 
\section{Collective--Matrix Structure}\label{sec:cmf} 
In this section we introduce equivalent representations  for the collective--matrix structure  and develop basic algebra for analyzing and manipulating collective--matrices. 
\subsection{Basic Notations}
Matrices are denoted by uppercase letters, $X$, $M$, etc. 
Matrix inner product is given by $\langle X,Y\rangle=\sum_{(i,j)}X_{ij}Y_{ij}$. The set of symmetric matrices of dimension $N$ is denoted as $\mathbb{S}^N$.  
For $M\in\mathbb{R}^{m\times n}$, with singular values  $\sigma_1\ge\sigma_2\ge\ldots$, common matrix norms include the \textit{nuclear norm} $\|M\|_*=\sum_{i}\sigma_i$, the \textit{spectral norm} $\|M\|_2=\sigma_1$, and  the \textit{Frobenius norm} $\|M\|_F=\sqrt{\sum_i\sigma_i^2}=\sqrt{\sum_{ij}M_{ij}^2}$.

\begin{definition}[Dual Norm]
Given any  norm $\|\cdot\|$ defined on a metric space $\mathcal{V}$, the \textit{dual norm}, $\|\cdot\|^{*}$ defined on dual space $\mathcal{V^*}$ is given by $\|X\|^{*}={\text{sup}}_{\|Y\|\le 1}\langle X,Y\rangle$.
\end{definition}

\begin{definition}[Operator Norm] Given a linear operator $\mathcal{P}:\mathcal{V}\to\mathcal{W}$, the operator norm of $\mathcal{P}$ is given by $\|\mathcal{P}\|_{\text{op}}=\underset{X\in\mathcal{V}\setminus\{0\}}{\text{sup}}\frac{\|\mathcal{P}(X)\|_\mathcal{W}}{\|X\|_\mathcal{V}}$, where $\|.\|_{\mathcal{V}}$ and $\|.\|_\mathcal{W}$ are the Euclidean norms in the respective spaces. 
\end{definition}

\subsection{Collective--Matrix Representation}
A collective--matrix, denoted using script letters, $\mathcal{X}$, $\mathcal{M}$, etc., is a collection of affinity relations among a set of $K$ types of \textit{entities}; and is primarily represented as a list  of $V$ matrices, $\mathcal{X}=[X_v]_{v=1}^V=[X_v: v=1,2,\ldots,V]$. Each component matrix $X_v$, called a \emph{view}, is the affinity matrix between a pair of entity types denoted by ${r_v}$ (entity type along rows) and $c_v$ (entity type along columns). We only consider static undirected affinity relations, wherein, for a given pair of entity types $k_1,k_2\in\{1,2,\ldots K\}$, there is at most one affinity relation $X_v$ defined between $k_1$ and $k_2$.

The entity--relationship structure defining a collective--matrix is represented by an undirected  graph $\mathcal{G}$, with nodes denoting the $K$ entity types, and an  edge between nodes $k_1$ and $k_2$ implying that a  view $X_v$ with  either $(r_v\,=\,k_1,c_v=k_2)$ or $(r_v=k_2,c_v=k_1)$ exists in the collective matrix. We assume that the graph $\mathcal{G}$ forms a single connected component, if not, each connected component could be handled separately without loss of generality. 
An illustration of a collective matrix structure $\mathcal{X}$ and its entity--relationship graph $\mathcal{G}$ is given in Fig.~\ref{fig:example} $(a)$--$(b)$. 

For $k=1,2,\ldots, K$, denote the number of instances of the $k^{\text{th}}$ entity type by $n_k$; let $N=\sum_k n_k$. Then, $\forall v$, $X_v\in\mathbb{R}^{n_{r_v}\times n_{c_v}}$, and  collective--matrices with common entity--relationship graph $\mathcal{G}$ belong to the  space:
 $$\quad\mathfrak{X}=\mathbb{R}^{n_{r_1}\times n_{c_1}}\times \mathbb{R}^{n_{r_2}\times n_{c_2}}\times\ldots\times \mathbb{R}^{n_{r_V}\times n_{c_V}}.$$
Finally, $\forall v$, $\mathcal{I}(v)=\{(i,j):i\in[n_{r_v}], j\in[n_{c_v}]\}=[n_{r_v}]\times [n_{c_v}]$ denotes the set indices representing the elements in view $v$, where $[N]=\{1,2,\ldots,N\}$.

\subsubsection{Equivalent Representations}\label{sec:altRep}
\begin{figure}[t]
\centering
\includegraphics[width=\columnwidth]{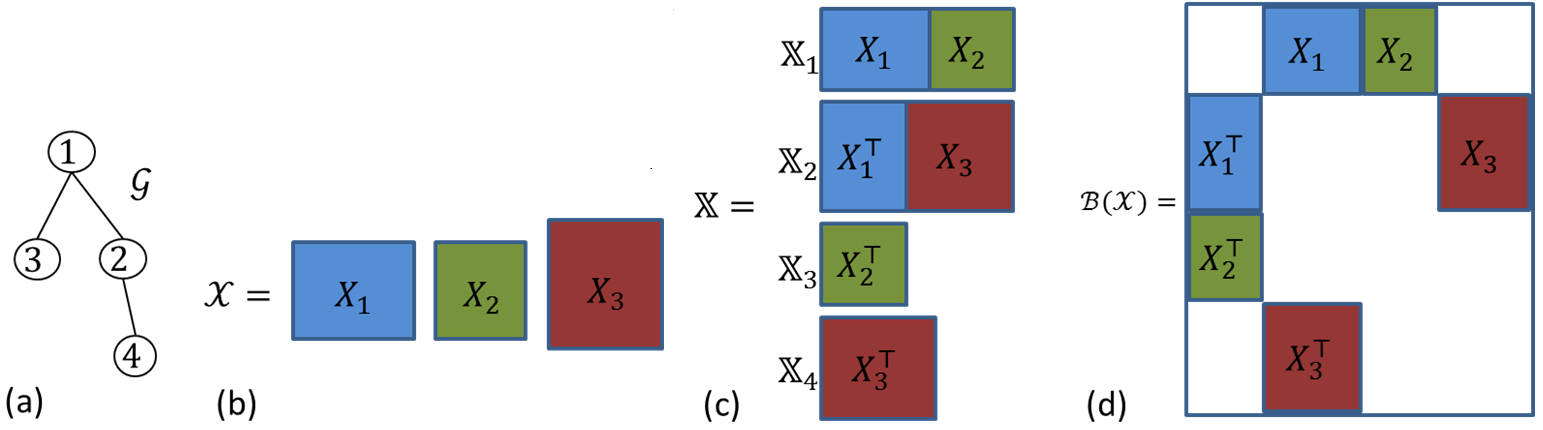}
\caption{An illustration of the various collective--matrix representations described in Section~\ref{sec:cmf}}
\label{fig:example}
\end{figure}

For mathematical convenience, we introduce two alternate (equivalent) representations for collective--matrices. These are used interchangeably in the rest of the paper.
\begin{asparaenum}
\item  \textbf{Entity Matrix Set Representation:} A collective--matrix $\mathcal{X}$, can be equivalently represented as a set of $K$ matrices $\mathbb{X}=[\mathbb{X}_k]_{k=1}^K$, such that $\mathbb{X}_k$ is a matrix formed by concatenating (appropriately transposed) views  involving the entity type $k$. Let $\mathbbm{1}_E$ denote the indicator variable for statement $E$, and the operator $\text{hcat}\{\}$ denote horizontal concatenation of a list. We then have the column dimension of $\mathbb{X}_k$ given by $m_k=\sum_{v=1}^V n_{c_v}\mathbbm{1}_{(r_v=k)}+n_{r_v}\mathbbm{1}_{(c_v=k)}$, and \[\mathbb{X}_k:=\text{hcat}\big\{[X_v\mathbbm{1}_{(r_v=k)},X_v^\top\mathbbm{1}_{(c_v=k)}]_{v=1}^V\big\}\in\mathbb{R}^{n_k\times m_k}.\]
\item \textbf{Block Matrix Representation:} Collective--matrices can also be represented as blocks in a symmetric matrix of size $N\times N$, where $N=\sum_k n_k$ \cite{bouchard2013convex}. For a symmetric matrix $Z\in\mathbb{S}^{N}$, we identify $K\times K$ blocks, wherein the $(k_1,k_2)$ block, denoted as $Z[k_1,k_2]$, is of dimension ${n_{k_1}\times n_{k_2}}$. Block matrix representation for $\mathcal{X}$ is given by:
\[ \mathcal{B}(\mathcal{X})[k_1,k_2]= \left\{ \begin{array}{ll}X_v& \text{if } \exists v, \text{ s.t. } r_v=k_1,c_v=k_2\\X_v^\top& \text{if } \exists v, \text{ s.t. } r_v=k_2,c_v=k_1\\0& \text{otherwise}.\end{array}\right.\]
We define operators $P_v:S^N\to \mathbb{R}^{n_{r_v}\times n_{c_v}}$, such that $P_v({Z})={Z}[r_v,c_v]$; and $\forall {Z}\in\mathbb{S}^N$, $\mathcal{Z}=[P_v({Z})]_{v=1}^V\in\mathfrak{X}$.

These alternate representations for collective--matrix structure are illustrated in Figure~\ref{fig:example} $(c)$ and $(d)$, respectively.
\end{asparaenum}

\subsection{Collective--Matrix Algebra}\label{sec:algebra}
\begin{asparaenum}
\item [\textbf{Collective--Matrix Inner Product and Euclidean Norm}]
\begin{equation*}
\langle \mathcal{X},\mathcal{Y}\rangle=\sum_{v=1}^V\langle X_v,Y_v\rangle, \text{ and } \|\mathcal{X}\|_{F}=\sqrt{\langle \mathcal{X},\mathcal{X}\rangle}.
\end{equation*}
\textbf{Note:} We overload the notation for inner product $\langle\cdot,\cdot\rangle$, and the Frobenius norm $\|\cdot\|_F$ for matrices and collective--matrices, with operands providing disambiguation.
\item[\textbf{Standard Orthonormal Basis}]
The \textit{standard orthonormal basis} for $\mathfrak{X}$ is given by  $\{\mathcal{E}^{(v,i_v,j_v)}: v\in[V], (i_v,j_v)\in\mathcal{I}(v)\}$, where $\mathcal{E}^{(v,i_v,j_v)}\in\mathfrak{X}$ has a value of $1$ in the $(i_v,j_v)^\text{th}$ element of view $v$, and $0$ everywhere else. Recall that $[n]=\{1,2,\ldots, n\}$, and $\mathcal{I}(v)=[n_{r_v}]\times[n_{c_v}]$. 

\item[\textbf{Joint Factorization and Collective--Matrix Rank}] A collective-matrix $\mathcal{X}\in\mathfrak{X}$ is said to possess an \textit{$R$--dimensional joint factorization}, if there exists a set of factors $\{U_k\in\mathbb{R}^{n_k\times R}\}_{k=1}^{K}$, such that $\forall v,\;X_v=U_{r_v}U_{c_v}^\top$. The set of collective--matrices in $\mathfrak{X}$ that have a joint factorization structure of finite dimension is denoted by $\bar{\mathfrak{X}}\subseteq\mathfrak{X}$. 
For $\mathcal{X}\in\bar{\mathfrak{X}}$, the \textit{collective--matrix rank} is defined as the minimum value of $R$ such that an $R$--dimensional joint factorization exists for $\mathcal{X}$. 
\end{asparaenum}

\subsection{Atomic Decomposition of Collective--Matrices} \label{sec:atomicdecomp}
Consider the following set of rank--$1$ collective--matrices:
\begin{flalign}
\mathscr{A}=\text{ext}(\text{conv}\{[P_v(uu^\top)]_{v=1}^V:u\in\mathbb{R}^N, \|u\|_2=1\}),
\label{eq:atom}
\end{flalign}
where $\text{conv}()$ and $\text{ext}()$ return the convex hull and the extreme points of a set, respectively. Recall that $N=\sum_k n_k$, and $P_v:\mathbb{S}^N\to \mathbb{R}^{n_{r_v}\times n_{c_v}}$ extracts the block corresponding to the view $v$ in an $N\times N$ symmetric matrix. 
From the block matrix representation (Sec.~\ref{sec:altRep}), note that $\mathfrak{X}=\text{aff}(\mathscr{A})$; and the following proposition can be easily verified:
\begin{proposition} A collective--matrix has a joint factorization structure if and only if it belongs to the conic hull of $\mathscr{A}$, i.e. $\bar{\mathfrak{X}}=\text{cone}(\mathscr{A})$. \hfill\(\Box\)
\end{proposition}
We define the following quantities of interest:
\begin{compactitem}
\item [\textbf{Collective--Matrix Atomic Norm}:] also the gauge of $\mathscr{A}$,
\begin{flalign}
&\|\mathcal{X}\|_\mathscr{A}:=\text{inf}\{t>0: \mathcal{X}\in t\cdot \text{conv}(\mathscr{A})\}.&
\label{eq:atomicNorm}
\end{flalign}
\item [\textbf{Support function} of $\mathscr{A}$:]
\begin{flalign}
&\|\mathcal{X}\|_\mathscr{A}^*:=\text{sup}\{\langle \mathcal{X},\mathcal{A}\rangle: \mathcal{A}\in \mathscr{A}\}.&
\label{eq:dualNorm}
\end{flalign}
\item [\textbf{``sign" collective--matrices} of $\mathcal{X}$:]
\begin{flalign}
&\mathscr{E}(\mathcal{X})=\{\mathcal{E}:\|\mathcal{X}\|_\mathscr{A}=\langle\mathcal{E},\mathcal{X}\rangle, \|\mathcal{E}\|_\mathscr{A}^*=1\}.
\label{eq:zxex}
\end{flalign}
\end{compactitem}

\textbf{Remarks} 
\begin{asparaenum}
\item $\|\mathcal{X}\|_\mathscr{A}$ is not always a norm. It is a norm if $\mathscr{A}$ is centrally symmetric, i.e. if $ \mathcal{A}\in \mathscr{A}\Leftrightarrow -\mathcal{A}\in \mathscr{A}$. 
\item By convention, $\|\mathcal{X}\|_\mathscr{A}=\infty$ if $\mathcal{X}\in\mathfrak{X}\setminus\bar{\mathfrak{X}}$. 
\item  However, $\|\mathcal{X}\|_\mathscr{A}$ is always a convex function and  exhibits many norm--like properties. $\forall \mathcal{X}\in\mathfrak{X}$, $\|\mathcal{X}\|_\mathscr{A}\ge 0$ and $\|\mathcal{X}\|_\mathscr{A}=0$ iff $\mathcal{X}=0$; $\forall a\ge0,\; \|a\mathcal{X}\|_\mathscr{A}=a\|\mathcal{X}\|_\mathscr{A}$; and $\|\mathcal{X}+\mathcal{Y}\|_\mathscr{A}\le\|\mathcal{X}\|_\mathscr{A}+\|\mathcal{Y}\|_\mathscr{A}$.
\item If $\|\mathcal{X}\|_\mathscr{A}$ is a norm, then $\|\mathcal{X}\|_\mathscr{A}^*$ is its dual norm.
\end{asparaenum}

\subsubsection{Primal Dual representation}
For all $\mathcal{X}\in\bar{\mathfrak{X}}$, $\|\mathcal{X}\|_\mathscr{A}<\infty$, and the atomic norm defined in \eqref{eq:atomicNorm}, can be equivalently defined using the following primal and dual optimization problems. 
\begin{flalign}
(P) &\;\;\|\mathcal{X}\|_\mathscr{A}=\min_{\{\lambda_r\ge 0\}} \;\textstyle\sum_r \lambda_r \;\;\text{s.t. }\textstyle \sum_r\lambda_r\mathcal{A}_r=\mathcal{X},&\\
(D) &\;\;\|\mathcal{X}\|_\mathscr{A}=\max_{\mathcal{Y}\in \mathfrak{X}}\; \langle \mathcal{X},\mathcal{Y}\rangle \;\;\text{s.t. } \|\mathcal{Y}\|_\mathscr{A}^*\le 1.&
\end{flalign}
\begin{proposition} 
$\forall\mathcal{X}\in\bar{\mathfrak{X}}$,  convex programs  $(P)$ and $(D)$ defined above are equivalent to:
\begin{flalign*}
(P)&\;\;\|\mathcal{X}\|_\mathscr{A}=\min_{Z\in\mathbb{S}^N} \text{tr}(Z)\quad \text{s.t. }P_v[Z]=X_v \forall \;v,&\\
(D)&\;\;\|\mathcal{X}\|_\mathscr{A}=\max_{\mathcal{Y}\in \mathfrak{X}} \langle \mathcal{X},\mathcal{Y}\rangle \quad \text{s.t. } \frac{1}{2}\mathcal{B}(\mathcal{Y})\preccurlyeq \mathbb{I}.&
\end{flalign*}

\label{prop:atomicNormSDP}
\end{proposition}
\section{Convex Collective--Matrix Completion}\label{sec:convexEstimator}
Denote the  ground truth collective--matrix as $\mathcal{M}\in\bar{\mathfrak{X}}$. The task in collective--matrix completion is to recover $\mathcal{M}$ from a subset of the (potentially noisy) entries of $\mathcal{M}$. Denote the indices of observed entries by $\Omega=\{(v_s,i_s,j_s):(i_s,j_s)\in\mathcal{I}(v_s), s=1,2,\ldots, |\Omega|\}$. For conciseness, we denote the standard basis corresponding to indices in $\Omega$ as $\forall s$, $\mathcal{E}^{(s)}=\mathcal{E}^{(v_s,i_s,j_s)}$. Further, we define the operator $P_\Omega$ as:
\begin{equation}\textstyle{P_\Omega(\mathcal{X})=\sum_{s=1}^{|\Omega|}\langle \mathcal{X},\mathcal{E}^{(s)}\rangle\mathcal{E}^{(s)}.}
\label{eq:pomega}
\end{equation}
We consider two observation models:
\begin{asparaenum}
\item Noise--free model: $\mathcal{M}$ is observed on $\Omega$ without any noise, i.e.~$\forall s, y_s=\langle \mathcal{M},\mathcal{E}^{(s)}\rangle$.
\item Additive noise model: Entries of $\mathcal{M}$ on $\Omega$  are observed with additive random noise, i.e. $\forall s, y_s=\langle \mathcal{M},\mathcal{E}^{(s)}\rangle+\eta_s$.
\end{asparaenum}

\subsection{Assumptions}\label{sec:Ass}
Collective--matrix completion  is in general an ill--posed problem. However, recent literature on related tasks of compressed sensing \cite{donoho2006compressed,candes2006robust,candes2006near},  matrix estimation~\cite{recht2010guaranteed,candes2009exact,candes2010matrix,keshavan2010matrix,keshavan2010noise,negahban2012restricted,jain2013low,gunasekar2014exponential}, and other high dimensional estimation~\cite{negahban2009unified,chandrasekaran2012convex,candesmathematics,vershynin2014estimation} etc. propose tractable estimators  with strong statistical guarantees for such high dimensional problems when low dimensional structural constraints are imposed on the ground truth parameters. 

\begin{assumption}[$R$--dimensional joint factorization] We assume that the ground truth collective--matrix $\mathcal{M}$ has a collective--matrix rank of $R\ll N$, i.e. $\exists \{U_k\in\mathbb{R}^{n_k\times R}\}$, such that $\forall v$, $M_v=U_{r_v}U_{c_v}^\top$.\hfill\(\Box\) 
\label{ass:lr}
\end{assumption}

Analogous to matrices, $\forall \mathcal{X}\in\bar{\mathfrak{X}}$, we define the following: 
\begin{flalign}
 T(\mathcal{X})=&\text{aff}\{\mathcal{Y}\in\bar{\mathfrak{X}}:\forall\;v,  \text{rowSpan}(\mathbb{Y}_{r_v})\subseteq \text{rowSpan}(\mathbb{X}_{r_v})\text{\emph{ or }}\text{rowSpan}(\mathbb{Y}_{c_v})\subseteq \text{rowSpan}(\mathbb{X}_{c_v})\}, \label{eq:T}\\
 T^\perp(\mathcal{X})=&\{\mathcal{Y}\in\bar{\mathfrak{X}}:\forall\;v, \text{rowSpan}(Y_v)\perp \text{rowSpan}(X_v)\text{\emph{ and }}\text{colSpan}(Y_v)\perp \text{colSpan}(X_v)\},&
\label{eq:Tperp}
\end{flalign}
where we have used the entity matrix set representation in \eqref{eq:T}~(See Sec.~\ref{sec:altRep}).
In the rest of the paper, we denote $T(\mathcal{M})$ and $T^\perp(\mathcal{M})$ simply as $T$ and $T^\perp$, respectively. Let $P_T$ and $P_{T^\perp}$ be projections onto $T$ and $T^\perp$, respectively. 

\begin{lemma} $\forall \mathcal{X}\in\bar{\mathfrak{X}}$, $\mathcal{X}\in T^\perp$ iff $\langle \mathcal{X},\mathcal{Y}\rangle=0$, $\forall\mathcal{Y}\in T$. 
\\
\normalfont The lemma is proved in the supplementary material. \hfill\(\Box\)
\label{lem:t}
\end{lemma}	
As with matrix completion, in a localized  observation setting, consistent recovery is infeasible  if any entry in $\mathcal{M}$ is overly significant. 
Such cases are precluded through the following  analogue of \textit{incoherence conditions} \cite{candes2009exact,gross2011recovering}. 
\begin{assumption} [Incoherence] We assume that $\exists\;(\mu_0, \mu_1)$ such that the following incoherence conditions with respect to standard basis are satisfied for all $\mathcal{E}^{(v,i,j)}$:
\begin{flalign}
\label {eq:pt} &\|P_T(\mathcal{E}^{(v,i,j)})\|_{F}^2\le\frac{\mu_0R}{m_{r_v}}+\frac{\mu_0R}{m_{c_v}},&\\
&\exists \mathcal{E}_\mathcal{M} \in \mathscr{E}(\mathcal{M})\cap T\text{, s.t. }\langle\mathcal{E}^{(v,i,j)},\mathcal{E}_\mathcal{M}\rangle^2 \le \frac{\mu_1R}{N^2}.\;&
\end{flalign}
Recall $\mathscr{E}(\mathcal{M})$ from \eqref{eq:zxex}, and $m_k=\sum_{v=1}^V n_{c_v}\mathbbm{1}_{(r_v=k)}+n_{r_v}\mathbbm{1}_{(c_v=k)}$.

\normalfont Note that $\|P_T(\mathcal{E}^{(v,i,j)})\|_F^2$ is upper bounded by a sum of norms of  projections of $m_{r_v}$ and $m_{c_v}$ dimensional standard basis (in $\mathbb{R}^{m_{r_v}}$  and $\mathbb{R}^{m_{c_v}}$, respectively) onto the $R$ dimensional latent factor space. Equation~\eqref{eq:pt} ensures that no single latent dimension is overly dominant. \hfill\(\Box\)
\label{ass:incoh}
\end{assumption}

Further, in Section~\ref{sec:algebra} it was noted that in general $\bar{\mathfrak{X}}\subseteq{\mathfrak{X}}$, and the set of atoms spanning $\bar{\mathfrak{X}}$ defined in \eqref{eq:atom} need not be centrally symmetric. This poses subtle challenges in analyzing the consistency of collective--matrix completion. To mitigate these difficulties, we consider a restricted set of collective--matrix structures, under which $\mathfrak{X}=\bar{\mathfrak{X}}$, and $\mathscr{A}$ is centrally symmetric. 

\begin{assumption} [Bipartite $\mathcal{G}$] Recall from Section~\ref{sec:cmf} that the entity--relationship structure of $\mathfrak{X}$ is represented through an undirected graph $\mathcal{G}$. We assume that $\mathcal{G}$ is bipartite, or equivalently $\mathcal{G}$ does not contain any odd length cycles.

\normalfont Using induction, it can be easily verified that Assumption~\ref{ass:g} implies that $\mathfrak{X}=\bar{\mathfrak{X}}$, and that $\mathscr{A}$ is centrally symmetric. Under this assumption, $\|.\|_\mathscr{A}$ and $\|.\|_\mathscr{A}^*$ are norms, and $\|\mathcal{X}\|_\mathscr{A}^*=\frac{1}{2}\lambda_{\text{max}}(\mathcal{B}(\mathcal{X}))\le\frac{1}{2}\|\mathcal{B}(\mathcal{X})\|_2$. 
 We also note that for the well--posedness of collective--matrix completion, some variation of Assumptions~\ref{ass:lr}, and \ref{ass:incoh} is necessary. However, it is not clear if Assumption~\ref{ass:g} is necessary. \hfill\(\Box\)
\label{ass:g}
\end{assumption}

$\forall k$, we define $\Omega_k=\{(v_s,i_s,j_s)\in\Omega: r_{v_s}=k\text{ or }c_{v_s} \}$. Let $|\Omega_k|$ be the expected number of observations in $\Omega_k$. 
\begin{assumption}[Sampling] 
For $s\in[|\Omega|]$, independently\\(a) sample $k_s: k_s=k \text{ w.p. }\frac{|\Omega_k|}{2|\Omega|}$;\\(b)  sample $i_{k_s}\sim \text{uniform}([n_k])$; and\\(c) sample  $j_{k_s}\sim \text{uniform}([m_k])$. \\
$(v_s,i_s,j_s)$ is the index of  $(i_{k_s},j_{k_s})$ element in $\mathbb{M}_{k_s}$.   

\normalfont Given $v\in[V]$ and $(i,j)\in\mathcal{I}(v)$, and $s=1,2,\ldots,|\Omega|$:
\begin{equation}
\text{Pr}\big((v,i,j)=\Omega_s\big)=\frac{|\Omega_{r_v}|}{2|\Omega|n_{r_v}m_{r_v}}+\frac{|\Omega_{c_v}|}{2|\Omega|n_{c_v}m_{c_v}}.
\end{equation}
\textbf{Remarks:}
\begin{asparaenum}
\item Note that we overload the notation for cardinality of the set. $|\Omega_k|$ in the sampling scheme is the expected cardinality of $\Omega_k$, not the true cardinality of $\Omega_k$. However, Hoeffdings's inequality can be used to show that the cardinality of $\Omega_k$ concentrates sharply around the expectation, $|\Omega_k|$.
\item \textit{Why $|\Omega_k|$?}: For consistent recovery of $\mathcal{M}$, the low dimensional factors of $\mathcal{M}$, $\{U_k\in\mathbb{R}^{n_k\times R}\}$ need to be learnt. Given $k$, information on $U_k$ is entirely contained in $\mathbb{M}_k$. Thus, the optimal sample complexity for consistent recovery depends on  individual $|\Omega_k|$. The assumed sampling scheme is convenient for deriving bounds in terms of $|\Omega_k|$. 
\end{asparaenum}
\label{ass:sampling}
\end{assumption}

\subsection{Atomic Norm Minimization}
Collective--matrix rank of $\mathcal{M}\in\bar{\mathfrak{X}}$ is given by:
\[\text{rank}(\mathcal{M})=\min_{\{\lambda_r\ge0\}}\textstyle\sum_r \mathbbm{1}_{\lambda_r\neq 0}\quad s.t. \textstyle \sum_r\lambda_r\mathcal{A}_r=\mathcal{M},\] 
where $\mathcal{A}_r\in\mathscr{A}$. However, minimizing the rank of a collective--matrix is intractable. We use the atomic norm \eqref{eq:atomicNorm} as a convex surrogate for the rank function and propose the following convex estimator for the noise--free model:
\begin{equation}
\hat{\mathcal{M}}=\underset{\mathcal{X}\in\bar{\mathfrak{X}}}{\text{argmin}}\; \|\mathcal{X}\|_\mathscr{A}\quad \text{s.t.}\; P_\Omega(\mathcal{X})=P_\Omega(\mathcal{M}). 
\label{eq:noisefreeEst}
\end{equation}
For the additive--noise model, we suitably modify the above convex program to propose three equivalent estimators:
\begin{flalign}
\hat{\mathcal{M}}&=\underset{\mathcal{X}\in\bar{\mathfrak{X}}}{\text{argmin}}\; \|\mathcal{X}\|_\mathscr{A}\;\;\text{s.t. }\|P_\Omega(\mathcal{X}-\mathcal{M})\|_{F}^2\le\omega^2, &\label{eq:noise1}\\
\hat{\mathcal{M}}&=\underset{\mathcal{X}\in\bar{\mathfrak{X}}}{\text{argmin}}\; \|P_\Omega(\mathcal{X}-\mathcal{M})\|_{F}^2\;\;\text{s.t. }\|\mathcal{X}\|_\mathscr{A} \le \eta,&\label{eq:noise2}\\
\hat{\mathcal{M}}&=\underset{\mathcal{X}\in\bar{\mathfrak{X}}}{\text{argmin}}\; \|P_\Omega(\mathcal{X}-\mathcal{M})\|_{F}^2 + \gamma\|\mathcal{X}\|_\mathscr{A}.&\label{eq:noise3}
\end{flalign}
The estimators are theoretically equivalent in the sense that for some combination of $\omega$, $t$, and $\gamma$ we obtain the same estimate from the three convex programs. In practice, the  parameters are set through cross validation, and the choice of a convex program for noisy collective--matrix completion is often made by the algorithmic considerations. 
\section{Main Results}\label{sec:main}
The main result of the paper states that under the assumptions stated in Sec.~\ref{sec:Ass}, the convex program in \eqref{eq:noisefreeEst}, exactly recovers the ground truth collective--matrix with high probability. We then propose a scalable greedy algorithm  with convergence guarantees for solving noisy collective--matrix completion using \eqref{eq:noise2}.

\subsection{Consistency under Noise--Free Model}\label{sec:theory} 
Recall: $|\Omega_k|$ is the expected cardinality of $\Omega_k=\{(v,i,j)\in\Omega:r_v=k\text{ or }c_v=k\}$, with the true cardinality concentrating sharply under the sampling scheme (Assumption~\ref{ass:sampling}), and $|\Omega|$ is the cardinality of $\Omega$; $n_k$ is the number of instances of type $k$, and $N=\sum_k n_k$; $R$ is the collective--matrix rank of $\mathcal{M}$; and $\mu_0$ and $\mu_1$ are the incoherence parameters (Assumption~\ref{ass:incoh}). 

\begin{theorem} Assume that the following sample complexity requirements are met, 
\begin{compactenum}[(i)]
\item $\forall k,\; |\Omega_k|>c_0\mu_0n_kR\beta\log{N}\log{(N\kappa_\Omega(N))}$,
\item $|\Omega|>c_1\max\{\mu_0,\mu_1\}NR\beta\log{N}\log{(N\kappa_\Omega(N))}$,
\item  $\forall k$, $\frac{|\Omega_k|}{n_km_k}\ge c\frac{|\Omega|}{N^2}$ for some constant $c$,
\end{compactenum}
where $\kappa_\Omega(N)=\frac{3|\Omega|\sqrt{\max_k\frac{|\Omega_k|}{n_k m_k}}}{{\min_k\frac{|\Omega_k|}{n_k m_k}}}$, which  scales at most as $N^4$ for general $\Omega$ and as $N^2$ under the above requirements. Then,  under the  assumptions in Sec.~\ref{sec:Ass}, for large enough $c_0$, and $c_1$, and $\beta>1$, and noise--free observation model, the  convex program in \eqref{eq:noisefreeEst} exactly recovers the true collective--matrix $\mathcal{M}$ with probability greater than $1-N^{1-\beta}-c_2N^{1-\beta}\log{(N\kappa_\Omega(N))}$ for some  constant $c_2$. 

\label{thm:main}
\end{theorem}
\subsection{Algorithm}\label{sec:algo}
Recently, Jaggi~et.~al.~\cite{jaggi2010simple} proposed a scalable approximate algorithm for solving  nuclear norm regularized matrix estimation, by adapting the approximate SDP solver of Hazan~\cite{hazan2008sparse}. We observe that using the alternate formulation of  collective--matrix atomic norm stated in Proposition~\ref{prop:atomicNormSDP}, the convex program for noisy collective--matrix completion in \eqref{eq:noise2} can be cast as the following SDP:
\begin{equation}
\underset{Z\succcurlyeq0,}{\text{min}}\sum_{v=1}^V\|P_{\Omega_v}(M_v-P_v(Z))\|_F^2\;\;\text{ s.t. }\text{tr}(Z)\le \eta,
\label{eq:ccmf2}
\end{equation}
where $\Omega_v=\{(v_s,i_s,j_s)\in\Omega: v_s=v\}$. Hazan's algorithm for solving \eqref{eq:ccmf2} is given in Algorithm~\ref{alg:alg}.

\begin{algorithm}[H]
\caption{Hazan's Algorithm for Convex Collective--Matrix Completion \eqref{eq:ccmf2} (Hazans--CMC)}
\label{alg:alg}
\begin{algorithmic}
\State Rescale loss: $\hat{f}_\eta(Z)=\sum_{v}\|P_{\Omega_v}(M_v-P_v(\eta Z))\|_F^2$
\State Initialize $Z^{(1)}$
\ForAll {$t=1,2\ldots,T=\frac{4}{\epsilon}$}
\State Compute $u^{(t)}=\text{approxEV}\big(-\nabla\hat{f}_\eta(Z^{(t)}),\frac{1}{t^2}\big)$\footnote{$\text{approxEV}X,\epsilon\big)$ computes the approximate top eigen vector of $X$ upto $\epsilon$ error}
\State $\alpha_t:=\frac{2}{2+t}$
\State $Z^{(t+1)}=Z^{(t)}+\alpha_tu^{(t)}u^{(t)\top}$
\EndFor
\Return $[P_v(Z^{(T)})]_{v=1}^V$
\end{algorithmic}

\end{algorithm}
\begin{lemma}
Algorithm~\ref{alg:alg} returns an $\epsilon$ approximate solution to \eqref{eq:noise2} in time  $O\big(\frac{|\Omega|}{\epsilon^2}\big)$

\normalfont \textit{Proof:} From Theorem $2$ of Hazan's work~\cite{hazan2008sparse}, the proposed algorithm returns an  estimate for a SDP with primal--dual error of at most $\epsilon$ in $\frac{4C_f}{\epsilon}$ iterations, where $C_f$ is a curvature constant of the loss function. For squared loss, $C_f\le 1$ (Lemma 4 in \cite{jaggi2010simple}). 
Iteration $t$ in Algorithm~\ref{alg:alg} involves computing an $\frac{1}{t^2}$--approximate largest eigen value of a sparse matrix with $|\Omega|$ non--zero elements, which requires $O(\frac{|\Omega|}{t})$ computation using Lanczos algorithm.
\hfill\(\Box\)

\label{lem:algo}
\end{lemma}
In comparison, the SVT--CMC algorithm proposed by Bouchard~et.~al.~\cite{bouchard2013convex} converges faster in $O(\frac{1}{\sqrt{\epsilon}})$ iterations; however, each iteration in SVT--CMC requires computing all the non--zero eigen vectors of a $N\times N$ matrix, which does not scale well with $N$. Hazan's algorithm can be used to trade--off computation for accuracy in large datasets.

\subsection{Discussion and Directions for Future Work}
A collective--matrix $\mathcal{M}$ of collective--matrix rank $R$ lies in a lower dimensional model space spanned by the entity factors, $\{U_k\in\mathbb{R}^{n_k\times R}\}$. Given $k$, $U_k$ is estimated entirely from $P_{\Omega_k}(\mathbb{M}_k)$. Thus, an immediate lower bound on the sample complexity for well--posedness is given by $|\Omega_k|\sim O(n_k R)$. The results presented in the paper are optimal upto a poly--logarithmic factor.

A trivial estimate for collective--matrix completion is to estimate each component matrices independently. Since a joint low rank structure also imposes low rank structure on the component matrices, this is feasible if each component matrix satisfies the sample complexity requirements of standard matrix completion, i.e. $|\Omega_v|>C\max{\{\mu_0,\mu_1\}}R(n_{r_v}+n_{c_v})\log(n_{r_v}+n_{c_v})$. Another, estimate from standard matrix completion can be obtained by completing each matrix $\{\mathbb{M}_k\}$ in the entity--matrix set representation  independently, this requires a sample complexity of $|\Omega_k|>C\max{\{\mu_0,\mu_1\}}R(n_{k}+m_k)\log(n_k+m_k)$ for consistent recovery. In comparison to the sample complexity in Theorem~\ref{thm:main}, these results are sub--optimal as they do not completely leverage the shared structure introduced by the jointly factorizability of collective--matrices. 

Finally, the collective--matrix completion problem can also be cast as standard matrix completion problem of completing an incomplete $N\times N$ symmetric matrix, in which blocks corresponding to the collective--matrix are partially observed. However, the existing theoretical results on the consistency of matrix completion algorithms require either uniform random sampling  \cite{candes2009exact,keshavan2010matrix,jain2013low}, or coherent sampling \cite{chen2014coherent} of the entries of the matrix; and these results fail for blockwise random sampled matrix. Thus, our results provide a strict generalization to existing matrix completion results for the task of collective--matrix completion.

The key challenge in the analysis is to optimally leverage the shared structure. In high dimensional recovery, sample complexity depends on some complexity measure of the model space $T$. Compared to trivial extensions, $T$ defined in \eqref{eq:T} exploits the structure to give a narrow subspace for optimal sample complexity.

\section{Proof Sketch}\label{sec:proof}Detailed proofs of lemmata are included in the Appendix. The proof technique is analogous to the analysis for matrix completion. 

Let $\hat{\mathcal{M}}=\mathcal{M}+{\Delta}$ be the output of the convex program in \eqref{eq:noisefreeEst}. 
 The key steps in the proof are: 
\begin{asparaenum}
\item Show that under the sample complexity requirements of Theorem~\ref{thm:main}, $\|P_T(\Delta)\|_F$ can be upper bounded by a finite multiple of $\|P_{T^\perp}(\Delta)\|_F$. ($T$ and $T^\perp$ are defined in \eqref{eq:T}).
\item Show optimality of $\mathcal{M}$ for \eqref{eq:noisefreeEst} if a \textit{dual certificate} $\mathcal{Y}$ satisfying certain conditions exists.
\item Adapt the \textit{golfing  scheme}  introduced by Gross~et~al.~\cite{gross2011recovering} to construct $\mathcal{Y}$. 
\end{asparaenum}

We define $p(v,i,j)=\frac{|\Omega_{r_v}|}{2n_{r_v}m_{r_v}}+\frac{|\Omega_{c_v}|}{2n_{c_v}m_{c_v}}$, and note that for $s=1,2,\ldots,|\Omega|$, $\text{Pr}((v,i,j)=\Omega_s)=\frac{p(v,i,j)}{|\Omega|}$. We also define the following operators for $s=1,2,\ldots,|\Omega|$:
\begin{flalign}&\textstyle{
\mathcal{R}_s: \mathcal{X}\to \frac{1}{p(v_s,i_s,j_s)}\langle \mathcal{X},\mathcal{E}^{(s)}\rangle\;\mathcal{E}^{(s)} , \text{ and }}&\label{eq:rs}\\
&\textstyle{\mathcal{R}_\Omega: \mathcal{X}\to \sum_{s=1}^{|\Omega|}\mathcal{R}_s(\mathcal{X}) \text{ with }E[\mathcal{R}_\Omega]=\mathcal{I},}&\label{eq:romega}
\end{flalign}
where $\mathcal{I}$ is the identity operator, and $\mathcal{E}^{(s)}=\mathcal{E}^{(v_s,i_s,j_s)}$

\begin{lemma}
Let $\forall\; k$, $|\Omega_k|\ge c_0 \mu_0 n_kR\beta\log{N}$ for a large constant enough $c_0$. Then, under the assumptions in Sec.~\ref{sec:Ass},  the following holds w. p. greater than $1-N^{1-\beta}$,
\begin{equation*}\|P_T\mathcal{R}_\Omega P_T-P_T\|_\text{op}\le \frac{1}{2}.\end{equation*}
\normalfont Proof in the supplementary material.\hfill\(\Box\)
\label{lem:pt}
\end{lemma} 

Let $M_\Omega(v,i,j)$ denote the multiplicity of $(v,i,j)$ in $\Omega$, i.e. $M_\Omega(v,i,j)=\sum_s \mathbbm{1}_{(v,i,j)=(v_s,i_s,j_s)}$; we have  $M_\Omega(v,i,j)\le|\Omega|$. Also, note that $\min_k \frac{|\Omega_{k}|}{n_{k}m_{k}}\le p(v,i,j)\le \max_k \frac{|\Omega_{k}|}{n_{k}m_{k}}$. Thus, for all $\mathcal{X}$,
\begin{flalign}
 \|\mathcal{R}_\Omega(\mathcal{X})\|_F=\Big\|\sum_{\substack{v\in[V],\\(i,j)\in\mathcal{I}(v)}}\frac{M_\Omega(v,i,j)}{p(v,i,j)}\langle\mathcal{X},\mathcal{E}^{(v,i,j)}\rangle\mathcal{E}^{(v,i,j)}\Big\|_F\le  \frac{|\Omega|}{\min_k\frac{|\Omega_k|}{n_km_k}}\|\mathcal{X}\|_F,\label{eq:rpt}
\end{flalign}
Further,  using Lemma~\ref{lem:pt} we have the following w.h.p, 
\begin{flalign}
\nonumber \|\mathcal{R}_\Omega P_T(\Delta)\|^2_F&\ge \frac{1}{\max_k \frac{|\Omega_k|}{n_km_k}}\langle \mathcal{R}_\Omega P_T(\Delta), P_T(\Delta)\rangle = \frac{1}{\max_k \frac{|\Omega_k|}{n_km_k}}\langle P_T\mathcal{R}_\Omega P_T(\Delta), P_T(\Delta)\rangle&\\&\ge \frac{1}{2\max_k \frac{|\Omega_k|}{n_km_k}}\|P_T(\Delta)\|_F^2.
\label{eq:rptp}
\end{flalign}
 
Combining~\eqref{eq:rpt} and \eqref{eq:rptp}, along with $0=\|\mathcal{R}_\Omega(\Delta)\|_F\ge \|\mathcal{R}_\Omega P_{T}(\Delta)\|_F-\|\mathcal{R}_\Omega P_{T^\perp}(\Delta)\|_F$, 
\begin{equation}
\|P_T(\Delta)\|_F\le \frac{1}{2}\kappa_\Omega(N)\|P_{T^\perp}(\Delta)\|_F, 
\label{eq:result1}
\end{equation}
where $\kappa_\Omega(N)=\frac{3|\Omega|\sqrt{\max_k{|\Omega_k|}/{n_k m_k}}}{{\min_k{|\Omega_k|}/{n_k m_k}}}$.

\subsection{Optimality of $\mathcal{M}$}
\begin{lemma}
Under the assumptions in Sec.~\ref{sec:Ass}, let $\forall k$,  $|\Omega_k|\ge c_0 \mu_0 n_kR\beta\log{N}$ for a sufficiently large constant $c_0$. If there exists a dual certificate $\mathcal{Y}$ satisfying the following conditions, then $\mathcal{M}$ is the unique minimizer to \eqref{eq:noisefreeEst} w.p. greater than $1-N^{1-\beta}$:
\begin{compactenum}
\item $\|P_T(Y)-\mathcal{E}_\mathcal{M}\|_F\le \frac{1}{\kappa_\Omega(N)}$, and
\item $\|P_{T^\perp}(Y)\|_\mathscr{A}^*\le 1/2$,
\end{compactenum}
where recall $\mathcal{E}_\mathcal{M}$ from Assumption~\ref{ass:incoh}.\\\normalfont Proof is  in the supplementary material.\hfill\(\Box\)
\label{eq:optimality}

\label{lem:dual}
\end{lemma}
\subsection{Constructing Dual Certificate}
The proof is completed by constructing a dual certificate satisfying the conditions in Lemma~\ref{lem:dual}. We begin by partitioning each $\Omega$ into $p=\mathcal{O}(\log{(N\kappa_\Omega(N))})$ partitions denoted by $\Omega^{(j)}$, for $j=1,2,\ldots,p$, such that for all $j$: \\
(a) $\forall k$, $|\Omega^{(j)}_k|>c_0\mu_0\beta Rn_k\log{N}$ and $\frac{|\Omega_k^{(j)}|}{n_km_k}\le c\frac{|\Omega^{(j)}|}{N^2}$, \\
(b) $|\Omega^{(j)}|>c_2\max\{\mu_0,\mu_1\}\beta RN\log{N}$,\\ where $\Omega^{(j)}_k=\{(v,i,j)\in\Omega^{(j)}:r_v=k\text{ or }c_v=k\}$.


Define $\mathcal{W}_0=\mathcal{E}_\mathcal{M}$ where $\mathcal{E}_\mathcal{M}$ is the sign matrix from Assumption \ref{ass:incoh}. We define a process for $j=1,2,\ldots$ s.t. : 
\begin{equation}
\begin{aligned}
&\mathcal{Y}_j=\textstyle{\sum_{j^\prime=1}^j}\mathcal{R}_{\Omega^{(j^\prime)}}\mathcal{W}_{j^\prime-1}=\mathcal{R}_{\Omega^{(j)}}\mathcal{W}_{j-1}+\mathcal{Y}_{j-1},\\
&\mathcal{W}_j=\mathcal{E}_\mathcal{M}-P_T(\mathcal{Y}_j).
\end{aligned}
\end{equation}
Note that $\forall\;j$, $P_\Omega(\mathcal{Y}_j)=\mathcal{Y}_j$, and $P_T(\mathcal{W}_j)=\mathcal{W}_j$. We show that $\mathcal{Y}_p$ for $p= \mathcal{O}(\log{(N\kappa_\Omega(N))})$ satisfies the first condition required in Lemma~\ref{lem:dual}. The proof for second condition follows directly from the analogous proof for standard matrix completion by Recht~\cite{recht2011simpler} and is provided in the supplementary material.

It is easy to verify that $\frac{1}{2}\mathcal{E}^{(v,i,j)}\in\mathscr{A}$ for all ${(v,i,j)}$, and by Assumption~\ref{ass:g}, $-\frac{1}{2}\mathcal{E}^{(v,i,j)}\in\mathscr{A}$. Thus, $\forall \mathcal{X}\in\bar{\mathfrak{X}}$, 
\[\|\mathcal{X}\|_\mathscr{A}^*
\ge\frac{1}{2}\underset{\substack{v\in[V]\\(i,j)\in\mathcal{I}(v)}}{\text{max}}|\langle \mathcal{X},\mathcal{E}^{(v,i,j)}\rangle|\ge
\frac{1}{2N}\|\mathcal{X}\|_F.\] 
Also, $1=\|\mathcal{E}_\mathcal{M}\|_\mathscr{A}^*\ge\frac{1}{2N}\|\mathcal{E}_\mathcal{M}\|_F$, and  $P_T(\mathcal{Y}_p)-\mathcal{E}_\mathcal{M}=\mathcal{W}_p$. Using the above inequalities,  we have:
\begin{flalign}
 \textstyle{\|P_T(\mathcal{Y}_p)-\mathcal{E}_\mathcal{M}\|_F
=\|\mathcal{W}_{p-1}-P_T\mathcal{R}_{\Omega^{(p)}}\mathcal{W}_{p-1}\|_F}\overset{(a)}{\le}\frac{1}{2}\|\mathcal{W}_{p-1}\|_F\le\frac{1}{2^p}\|\mathcal{E}_\mathcal{M}\|_F\overset{(b)}{<}\frac{1}{\kappa_\Omega(N)}
\label{eq:yp}
\end{flalign} 
where $(a)$ follows from Lemma~\ref{lem:pt}, and $(b)$ follows for large enough $c_1$ s.t. $p=c_1\log{(N\kappa_\Omega(N))}$. Note that we use union bound to bound the probability of failure in $\mathcal{O}(\log{(N\kappa_\Omega(N))})$ partitions.
\section{Experiments}\label{sec:exp} 
\begin{figure}[htb]
\centering
\begin{subfigure}[b]{0.49\textwidth}
\includegraphics[width=\textwidth]{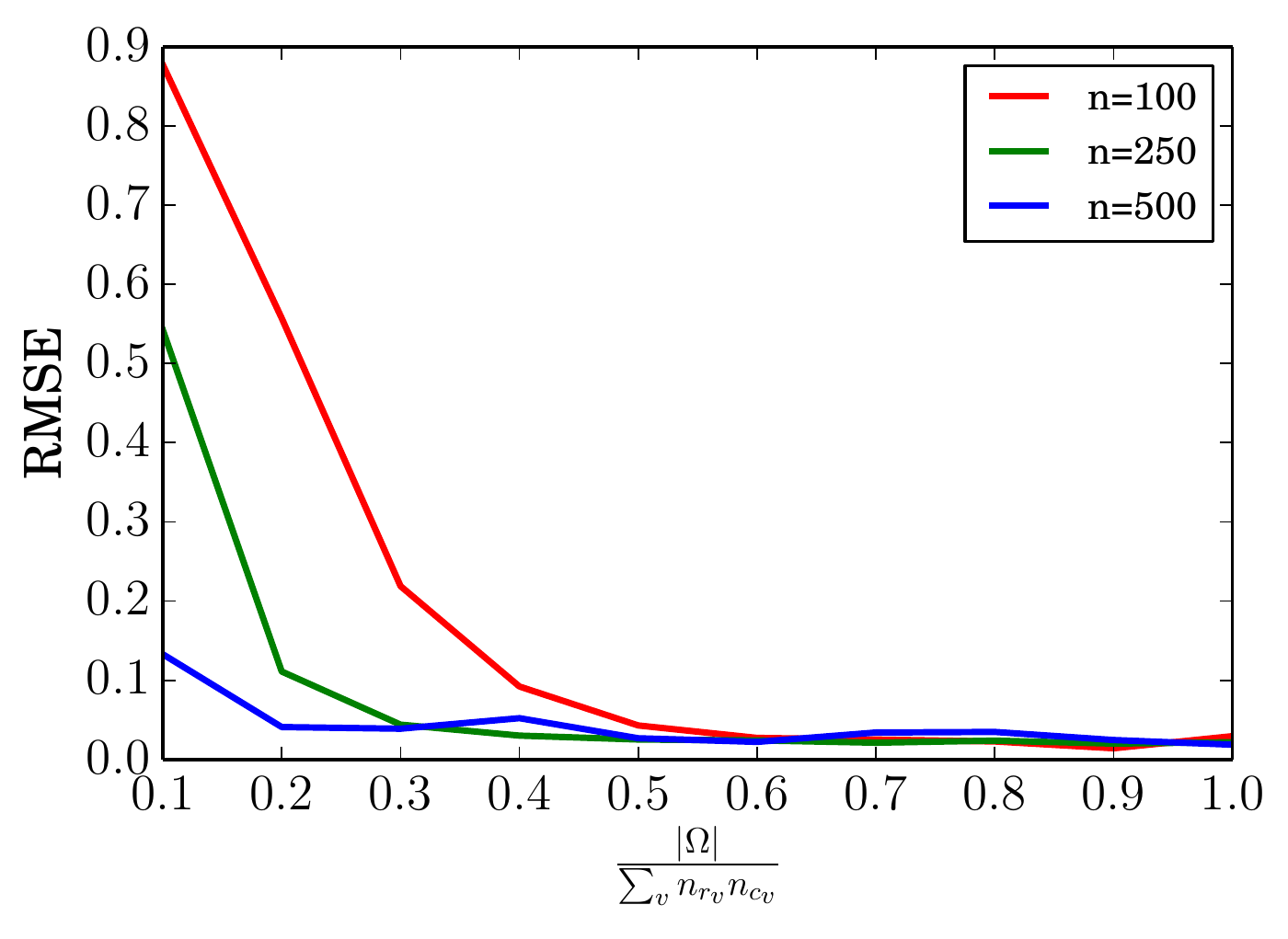}
\caption{RMSE vs unnormalized sample size
\label{fig:unnormalized}}
\end{subfigure}
~
\begin{subfigure}[b]{0.49\textwidth}
\includegraphics[width=\textwidth]{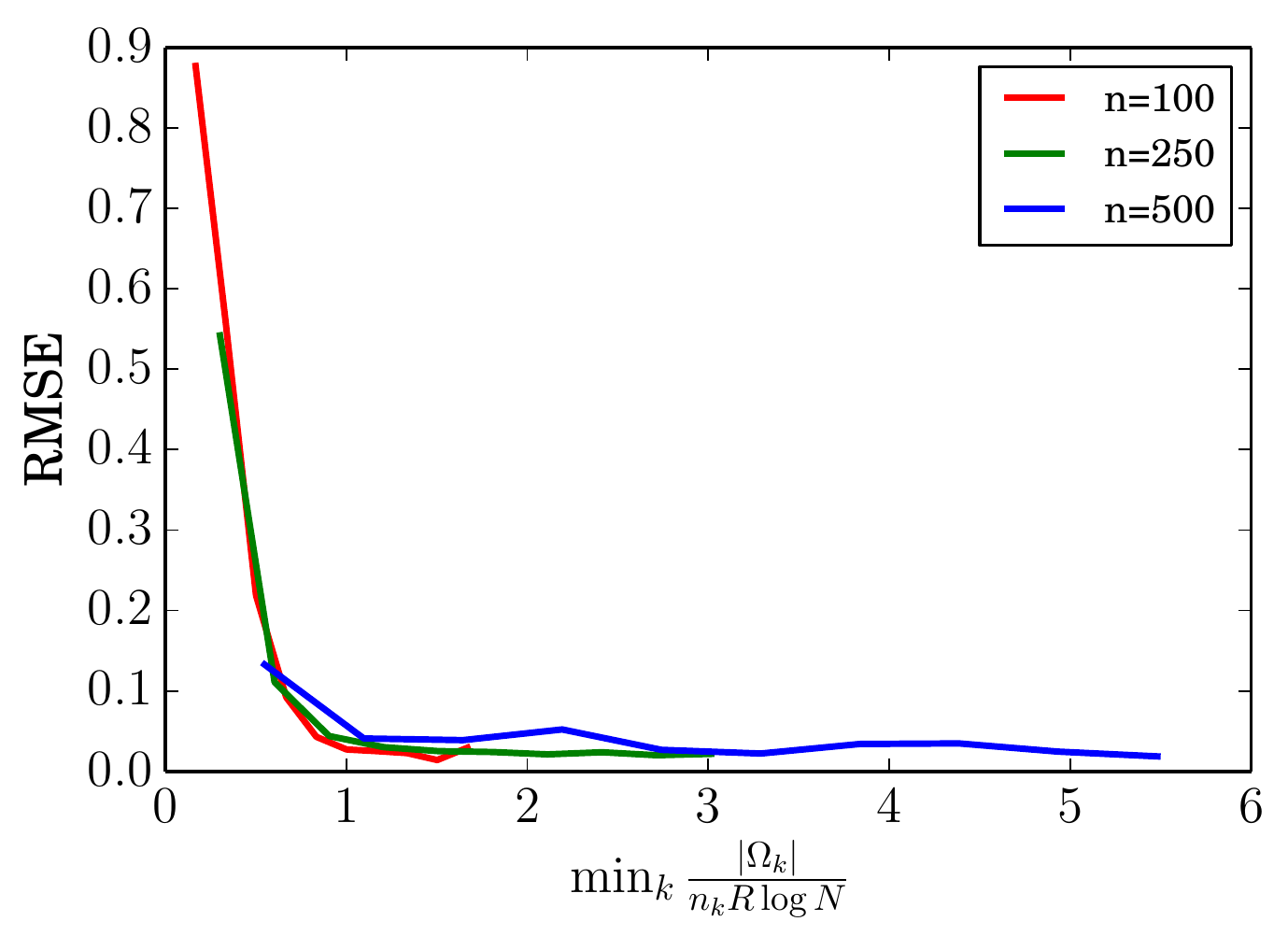}
\caption{RMSE vs normalized sample size
\label{fig:normalized}}
\end{subfigure}
\caption{Convergence of error measured against normalized and unnormalized sample size}
\end{figure}

The simulated experiments in this section are intended to corroborate our theoretical results in Sec.~\ref{sec:main}. 
We create low--rank ground truth collective--matrices with $K=4$, $V=3$, where view $1$ is a relation between entity types  $1$ and $2$, view $2$ is a relation between entity types $1$ and $3$, and view $3$ is a relation between entity types $2$ and $4$ respectively. For simplicity we assumed a common $n_k=n$. We create collective matrices with $n \in \{100,\;250,\;500\}$ and set the rank to $R=2\log{n}$. The matrices are partially observed with the fraction of observed entries, $\frac{|\Omega|}{\sum_v n_{r_v}n_{c_v}}$ varying as $[0.1,0.2,\ldots, 1]$. We plot the convergence of the errors against the unnormalized fraction of observations, $\frac{|\Omega|}{\sum_v n_{r_v}n_{c_v}}$ in Fig.~\ref{fig:unnormalized}, and against the normalized sample complexity provided by the theoretical analysis, $\min_k \frac{|\Omega_k|}{n_k R\log{N}}$ in Fig.~\ref{fig:normalized}. It can be seen from the plots that the error uniformly decays with increasing normalized sample size, indeed $|\Omega_k|>1.5n_kR\log{N}, \;\forall k$ samples suffice for the errors to decay to a very small value. The aligning of the curves (for different $n$)  given the  normalized sample size corroborates the theoretical sample complexity requirements.

\small
\bibliography{bibliography,latestbib}
\bibliographystyle{plain}
\appendix
\section{Operator Bernstein Inequality}\label{app:bernstein}

\begin{theorem}[Operator Bernstein Inequality~\cite{tropp2012user}] Let $S_i$, $i=1,2,\ldots,m$  be i.i.d self--adjoint operators of dimension $N$. If there exists constants $R$ and $\sigma^2$, such that $\forall i\; \|S_i\|_\text{op}\le R$ a.s., and $\sum_i\|E[S_i^2]\|_\text{op}\le\sigma^2$,
\begin{equation}
\text{then}\quad\textstyle \forall\; t>0\quad Pr\big(\|\sum_i S_i\|_\text{op}> t\big)\le N\exp{\Big(\frac{-t^2/2}{\sigma^2+\frac{Rt}{3}}\Big)}
\label{eq:bernstein}
\end{equation}

\end{theorem}

\section{Proof of Lemma~$\boldsymbol{1}$} 
Recall that:
\begin{compactitem}
\item $T(\mathcal{X})=\text{aff}\{\mathcal{Y}\in\bar{\mathfrak{X}}:\forall\;v,  \text{rowSpan}(\mathbb{Y}_{r_v})\subseteq \text{rowSpan}(\mathbb{X}_{r_v})\text{\emph{ or }}\text{rowSpan}(\mathbb{Y}_{c_v})\subseteq \text{rowSpan}(\mathbb{X}_{c_v})\}$
\item $T^\perp(\mathcal{X})=\{\mathcal{Y}\in\bar{\mathfrak{X}}:\forall\;v, \text{rowSpan}(Y_v)\perp \text{rowSpan}(M_v)\text{\emph{ and }}\text{colSpan}(Y_v)\perp \text{colSpan}(M_v)\}$
\end{compactitem}

We need to show that $\forall \mathcal{X}\in\bar{\mathfrak{X}}$, $\mathcal{X}\in T^\perp$ iff $\langle \mathcal{X},\mathcal{Y}\rangle=0$, $\forall\mathcal{Y}\in T$. 

$\Longrightarrow$ Let $\mathcal{X}\in\{\mathcal{X}\in\bar{\mathfrak{X}}:\langle \mathcal{X},\mathcal{Y}\rangle=0,\forall\mathcal{Y}\in T\}$, if $\mathcal{X}\notin T^\perp$, then $\exists v$ such that atleast one of the statements below hold true:
\begin{compactenum}[(a)]
\item $\text{rowSpan}(X_v)\not\perp \text{rowSpan}(M_v)$, or 
\item $\text{colSpan}(X_v)\not\perp \text{colSpan}(M_v)$
\end{compactenum}
WLOG let us assume that $(a)$ is true, the proof for the other case is analogous. Consider the decomposition $X_v=X_v^{(1)}+X_v^{(2)}$ such that $\text{rowSpan}(X_v^{(1)})\perp \text{rowSpan}(M_v)$ and $\text{rowSpan}(X_v^{(2)})\subseteq \text{rowSpan}(M_v)$. Consider the collective matrix $\mathcal{Y}$ such that $Y_{v^\prime}=X_v^{(2)}$  if $v^\prime=v$, and $Y_{v^\prime}=0$ otherwise. Clearly, $\mathcal{Y}\in T$ as $\forall\;v,  \text{rowSpan}(\mathbb{Y}_{r_v})\subseteq \text{rowSpan}(\mathbb{X}_{r_v})$, but $\langle \mathcal{X},\mathcal{Y}\rangle\neq 0$, a contradiction.

$\Longleftarrow$ If $\mathcal{X}\in T^\perp$, then by the definitions, $\forall \mathcal{Y}\in T$, $\langle \mathcal{X},\mathcal{Y}\rangle=\sum_v\langle X_v,Y_v\rangle=0$. 

\section{Proof of Lemma~$\boldsymbol{3}$}
Recall $\mathcal{R}_s$ and $\mathcal{R}_\Omega$ from $(18)$ and $(19)$. Also recall that $\forall \mathcal{X}\in\mathfrak{X},$ $ \mathcal{X}=\sum_{v=1}^V\sum_{(i,j)\in\mathcal{I}(v)}\langle \mathcal{X},\mathcal{E}^{(v,i,j)}\rangle\mathcal{E}^{(v,i,j)}$.

Thus, $P_T(\mathcal{X})=\sum_{v=1}^V\sum_{(i,j)\in\mathcal{I}(v)}\langle P_T(\mathcal{X}),\mathcal{E}^{(v,i,j)}\rangle\mathcal{E}^{(v,i,j)}=\sum_{v=1}^V\sum_{(i,j)\in\mathcal{I}(v)}\langle \mathcal{X},P_T(\mathcal{E}^{(v,i,j)})\rangle\mathcal{E}^{(v,i,j)}$

Define $\mathcal{V}_s:=P_T\mathcal{R}_sP_T: \mathcal{X}\to \frac{1}{p(v_s,i_s,j_s)}\langle \mathcal{X},P_T(\mathcal{E}^{(s)})\rangle P_T(\mathcal{E}^{(s)})$, where $p(v,i,j)=\frac{|\Omega_{r_v}|}{2n_{r_v}m_{r_v}}+\frac{|\Omega_{c_v}|}{2n_{c_v}m_{c_v}}$.

We then have $E[\mathcal{V}_s]=\frac{1}{|\Omega|}P_T$, and 
\begin{flalign}
\nonumber \|\mathcal{V}_s\|_\text{op}&=\underset{\|\mathcal{X}\|_F=1}{\text{sup}}\frac{1}{p(v_s,i_s,j_s)}\langle \mathcal{X},P_T(\mathcal{E}^{(s)})\rangle \|P_T(\mathcal{E}^{(s)})\|_F=\frac{1}{p(v_s,i_s,j_s)}\|P_T(\mathcal{E}^{(s)})\|^2_F&\\
&\overset{(a)}{\le}\frac{1}{p(v_s,i_s,j_s)}\left(\frac{\mu_0R}{m_{r_{v_s}}}+\frac{\mu_0R}{m_{c_{v_s}}}\right)\overset{(b)}{\le} \frac{1}{c_0\beta\log{N}},
\end{flalign}
where $(a)$ follows from the incoherence condition in Assumption $2$, and $(b)$ follows as $\forall k$, $|\Omega_k|>c_0\mu_0n_kR\beta\log{N}$. 
\begin{asparaenum}[(i)] \item Bound on $\|\mathcal{V}_s-E[\mathcal{V}_s]\|_\text{op}$
\begin{flalign}
&\|\mathcal{V}_s-E[\mathcal{V}_s]\|_\text{op}\overset{(a)}{\le}\max{(\|\mathcal{V}_s\|_\text{op}, \|E[\mathcal{V}_s]\|_\text{op})}\le\max{(\frac{1}{c_0\beta\log{N}},\frac{1}{\Omega})}=\frac{1}{c_0\beta\log{N}}
\end{flalign}
where $(a)$ follows as both $\mathcal{V}_s$ and $E[\mathcal{V}_s]$ are positive semidefinite. 
\item Bound on $\sum_{s=1}^{|\Omega|}\|E[(\mathcal{V}_s-E[\mathcal{V}_s])^2]\|_\text{op}$. \\
\begin{flalign}
\nonumber E[(\mathcal{V}_s)^2(X)]&=E\left[\frac{1}{p(v_s,i_s,j_s)^2}\langle \mathcal{X},P_T(\mathcal{E}^{(s)})\rangle \|P_T(\mathcal{E}^{(s)})\|_F^2P_T(\mathcal{E}^{(s)})\right]\\
&\le\frac{1}{c_0\beta\log{N}}E\left[\frac{1}{p(v_s,i_s,j_s)}\langle \mathcal{X},P_T(\mathcal{E}^{(s)})\rangle P_T(\mathcal{E}^{(s)})\right]=\frac{1}{|\Omega|c_0\beta\log{N}}P_T(\mathcal{X}).
\end{flalign}
\begin{equation}\|E[(\mathcal{V}_s-E[\mathcal{V}_s])^2]\|_\text{op}=\|E[\mathcal{V}_s^2]-(E[\mathcal{V}_s])^2]\|_\text{op}{\le}\max{(\|E[\mathcal{V}_s^2]\|_\text{op}, \|(E[\mathcal{V}_s])^2\|_\text{op})}\overset{(a)}{\le}\frac{1}{|\Omega|c_0\beta\log{N}},
\end{equation} 
where $(a)$ follows as $\|P_T\|_\text{op}\le 1$. 

Thus, $\sigma^2:=\sum_{s=1}^{|\Omega|}\|E[(\mathcal{V}_s-E[\mathcal{V}_s])^2]\|_\text{op}\le\frac{1}{c_0\beta\log{N}}$
\item  The lemma follows by using $(i)$ and $(ii)$ above in the operator Bernstein inequality in \eqref{eq:bernstein}.
\end{asparaenum}

\section{Proof of Lemma~$\boldsymbol{4}$}
Recall that under the assumptions made in the paper $\|\cdot\|_\mathscr{A}$ is norm, and by the sub differential characterization of norms we have the following:
\begin{equation}
\partial\|\mathcal{M}\|_\mathscr{A}=\{\mathcal{E}+\mathcal{W}:\mathcal{E}\in\mathscr{E}(\mathcal{M})\cap T, \mathcal{W}\in T^\perp, \|\mathcal{W}\|_\mathcal{A}^*\le1\}
\end{equation}
Recall $\mathscr{E}(\mathcal{M})$ from $(4)$. In particular the set $\{\mathcal{E}_\mathcal{M}+\mathcal{W}:\mathcal{W}\in T^\perp, \|\mathcal{W}\|_\mathcal{A}^*\le1\}\subset\partial\|\mathcal{M}\|_\mathscr{A}$, where $\mathcal{E}_\mathcal{M}$ is the sign vector from Assumption 2. 

Given any $\Delta,\text{with }P_\Omega(\Delta)=0$, consider any $\mathcal{W}\in T^\perp$, such that $\|P_{T^\perp}(\Delta)\|_\mathscr{A}=\langle \mathcal{W},P_{T^\perp}({\Delta})\rangle$ and $\mathcal{E}_\mathcal{M}+\mathcal{W}\in\partial\|\mathcal{M}\|_\mathscr{A}$. Let $\mathcal{Y}=P_\Omega(\mathcal{Y})$ be a dual certificate satisfying the conditions  stated in the Lemma.
\begin{flalign}
\nonumber \|\mathcal{M}+\Delta\|_\mathscr{A}&\overset{(a)}{\ge}\|\mathcal{M}\|_\mathscr{A}+\langle \mathcal{E}_\mathcal{M}+\mathcal{W}-\mathcal{Y},\Delta\rangle =\|\mathcal{M}\|_\mathscr{A}+\langle \mathcal{E}_\mathcal{M}-P_T(\mathcal{Y}),P_T(\Delta)\rangle +\langle \mathcal{W}-P_{T^\perp}(\mathcal{Y}),P_{T^\perp}(\Delta)\rangle &\\
\nonumber &\overset{(b)}{\ge}\|\mathcal{M}\|_\mathscr{A}-\|\mathcal{E}_\mathcal{M}-P_T(\mathcal{Y})\|_F\|P_T(\Delta)\|_F +\|P_{T^\perp}(\Delta)\|_\mathscr{A}(1-\|P_{T^\perp}(\mathcal{Y})\|_\mathscr{A}^*)&\\
&\overset{(c)}{\ge}\|\mathcal{M}\|_\mathscr{A}-\frac{1}{2}\kappa_\Omega(N)\|\mathcal{E}_\mathcal{M}-P_T(\mathcal{Y})\|_F\|P_{T^\perp}(\Delta)\|_{F}+\frac{1}{2}\|P_{T^\perp}(\Delta)\|_{\mathscr{A}}\overset{(d)}{>}\|\mathcal{M}\|_\mathscr{A},
\end{flalign}
where $(a)$ follows as $\langle \Delta,\mathcal{Y}\rangle =0$, $(b)$ follows from Holder's inequality, $(c)$ follows as  $\|P_{T^\perp}(\mathcal{Y})\|_\mathscr{A}^*\le\frac{1}{2}$ and $\frac{1}{2}\kappa_\Omega(N)\|P_{T^\perp}(\Delta)\|_{F}\ge\|P_{T}(\Delta)\|_{F}$ w.h.p. (from $(22)$), and $(d)$ follows as $\|\mathcal{E}_\mathcal{M}-P_T(\mathcal{Y})\|_F<\frac{1}{\kappa_\Omega(N)}$ and using $\|\mathcal{X}\|_\mathscr{A}=\min_{Z\succcurlyeq 0} tr(Z)\;\;s.t. P_v[Z]=X_v\;\forall v\ge \min_{Z\succcurlyeq 0} \|Z\|_F\;\;s.t. P_v[Z]=X_v\;\forall v\ge\|\mathcal{X}\|_F$.
\section{Dual Certificate--Bound on $\|P_{T^\perp}\mathcal{Y}_p\|_\mathscr{A}^*$}
Recall that  $\mathcal{Y}_p$ was constructed through a iterative process described in Sec.~$5.2$ following a golfing scheme introduced by Gross~et~al.~\cite{gross2011recovering}. The proof for the second property of the dual certificate, extends directly from the analogous proof for matrix completion by Recht~\cite{recht2011simpler}. We note that:
\begin{flalign}
\|P_{T^\perp}\mathcal{Y}_p\|_\mathscr{A}^*\le\sum_{j=1}^p\|P_{T^\perp}\mathcal{R}_{\Omega^{(j)}}\mathcal{W}_{j-1}\|_\mathscr{A}^*=\sum_{j=1}^p\|P_{T^\perp}(\mathcal{R}_{\Omega^{(j)}}-\mathcal{I})\mathcal{W}_{j-1}\|_\mathscr{A}^*\le \sum_{j=1}^p\|(\mathcal{R}_{\Omega^{(j)}}-\mathcal{I})\mathcal{W}_{j-1}\|_\mathscr{A}^*
\label{eqn:8}
\end{flalign}
Denote $\max_{(v,i,j)} |\langle \mathcal{X},\mathcal{E}^{(v,i,j)}\rangle|=\|\mathcal{X}\|_\text{max}$.

We state the following lemmas which are directly adapted from Theorem $3.5$ and Lemma $3.6$ in \cite{recht2011simpler}:
\begin{lemma}
Let $\Omega$ be any subset of entries of size $|\Omega|$ sampled independently according to Assumption~$4$, such that $E[\mathcal{R}_s(\mathcal{W})]=\frac{1}{|\Omega|}\mathcal{W}$, then for all $\beta>1$ and $N\ge2$, the following holds with probability greater than $1-N^{1-\beta}$ provided $|\Omega|>6N\beta \log{N}$, and $\frac{|\Omega_k|}{n_km_k}\ge \frac{|\Omega|}{N^2};\forall k$:
\begin{equation}
\|(\mathcal{R}_{\Omega}-\mathcal{I})\mathcal{W}\|_\mathscr{A}^*\le \|\mathcal{B}(\mathcal{R}_{\Omega}\mathcal{W}-\mathcal{W})\|_2\le \sqrt{\frac{8\beta N^3\log{N}}{3|\Omega|}}\|\mathcal{W}\|_\text{max}
\end{equation}
\end{lemma}
\emph{Proof.} The proof is obtained by applying the steps described for the analogous proof in ~\cite{recht2011simpler} on $\|\mathcal{B}(\mathcal{R}_{\Omega}\mathcal{W}-\mathcal{W})\|_2$. For $s=1,2,\ldots,|\Omega|$, let $\mathcal{V}_s=\mathcal{B}(\mathcal{R}_s(\mathcal{W}))$, then $\mathcal{B}(\mathcal{R}_{\Omega}\mathcal{W}-\mathcal{W})=\sum_{s=1}^{|\Omega|}(\mathcal{V}_s-E[\mathcal{V}_s])$ is a sum of independent zero mean random variables. From the proof of Theorem $3.5$ in the work by Recht~\cite{recht2010guaranteed}, we have that for any $N\times N$ matrix $Z$, $\|	Z\|_2\le N\|Z\|_\text{max}$. 
\begin{asparaenum}[(i)]
\item $\|\mathcal{V}_s-E[\mathcal{V}_s]\|_2\le\|\mathcal{V}_s\|_2+\|E[\mathcal{V}_s]\|_2\overset{(a)}{\le}\frac{N^2}{|\Omega|}\|\mathcal{W}\|_\text{max}+\frac{N}{|\Omega|}\|\mathcal{W}\|_\text{max}\le \frac{3N^2}{2|\Omega|}\|\mathcal{W}\|_\text{max}$ for $N\ge2$, where $(a)$ follows as $\frac{1}{p(v,i,j)}\le\frac{1}{\min_k \frac{|\Omega_k|}{n_km_k}}\le \frac{N^2}{|\Omega|}$ if $\frac{|\Omega_k|}{n_km_k}\ge \frac{|\Omega|}{N^2},\forall k$; and $\|E[\mathcal{V}_s]\|_2=\frac{1}{|\Omega|}\|\mathcal{B}(\mathcal{W})\|_2$.
\item $\|E[(\mathcal{V}_s-E[\mathcal{V}_s])^2]\|_2=\|E[\mathcal{V}_s^2]-(E[\mathcal{V}_s])^2\|_2\le\max{\{\|E[\mathcal{V}_s^2]\|_2,\|(E[\mathcal{V}_s])^2\|_2\}}$.  

Now, $\|(E[\mathcal{V}_s])^2\|_2=\frac{1}{|\Omega|^2}\|\mathcal{B}(\mathcal{W})*\mathcal{B}(\mathcal{W})\|_2\le\frac{N^2}{|\Omega|^2}\|\mathcal{W}\|_\text{max}^2$.\\ Also, $\|E[\mathcal{V}_s^2]\|_2=\frac{1}{|\Omega|}\Big\|\displaystyle{\sum_{v=1}^V\sum_{(i,j)\in\mathcal{I}(v)}\frac{1}{p(v,i,j)}\langle \mathcal{W},\mathcal{E}^{(v,i,j)}\rangle \mathcal{B}{(\mathcal{E}^{(v,i,j)})}}\Big\|_2\le \frac{N^4}{|\Omega|^2}\|\mathcal{W}\|_\text{max}^2$. \\
Thus $\sigma^2:=\|E[(\mathcal{V}_s-E[\mathcal{V}_s])^2]\|_2\le \frac{N^4}{|\Omega|^2}\|\mathcal{W}\|_\text{max}^2$
\end{asparaenum}
The proof follows by using the above bounds in operator Bernstein's inequality with $t=\sqrt{\frac{8\beta N^3\log{N}}{3|\Omega|}}\|\mathcal{W}\|_\text{max}$


\begin{lemma}
If $\forall k$, $|\Omega_k|\ge c_0\beta n_kR\log{N}$, and the  Assumptions in $3.1$ are satisfied, then for sufficiently large $c_0$, the following holds with probability greater that $1-N^{1-\beta}$:
\begin{equation}
\forall\;\mathcal{W}\in T\; \|P_T\mathcal{R}_{\Omega}\mathcal{W}-\mathcal{W}\|_\text{max} \le \frac{1}{2} \|\mathcal{W}\|_\text{max}\end{equation}
\end{lemma}

Using the above lemmas in $(32)$, we have:
\begin{flalign}
\nonumber\|P_{T^\perp}\mathcal{Y}_p\|_\mathscr{A}^*&\le\sum_{j=1}^p\|(\mathcal{R}_{\Omega^{(j)}}-\mathcal{I})\mathcal{W}_{j-1}\|_\mathscr{A}^*\overset{(a)}{\le}\sum_{j=1}^p\sqrt{\frac{8\beta N^3\log{N}}{3|\Omega^{(j)}|}}\|\mathcal{W}_{j-1}\|_\text{max}\\&\overset{(b)}{\le}2 \sum_{j=1}^p 2^{-j}\sqrt{\frac{8\beta N^3\log{N}}{3|\Omega^{(j)}|}}\|\mathcal{E}_\mathcal{M}\|_\text{max}\overset{(c)}{\le}2 \sum_{j=1}^p 2^{-j}\sqrt{\frac{8\beta \mu_1R N\log{N}}{3|\Omega^{(j)}|}}\overset{(d)}{\le} \frac{1}{2},
\end{flalign}
where $(a)$ follows from Lemma $5$, $(b)$ from Lemma $6$ as $\mathcal{W}_j=\mathcal{W}_{j-1}-P_T\mathcal{R}_\Omega\mathcal{W}_{j-1}$,  $(c)$ from the second incoherence condition in Assumption~$2$, and finally $(d)$ if for large enough $c_1$,  $|\Omega^{(j)}|> c_1\mu_1\beta RN\log{N}$.

Finally, the probability that the proposed dual certificate $\mathcal{Y}_p$ fails the conditions of Lemma~$4$ is given by a union bound of the failure probabilities of $(24)$, Lemma~$5$, and $6$  for any partition $\Omega^{(j)}$: $3c_1\log{(N\kappa_\Omega(N))}N^{1-\beta}$; thus proving Theorem~$1$. 

\subsection{Proof of Lemma 6}
Using union bound and noting that $\sum_{v} n_{r_v}n_{c_v}\le N^2$, we have:
\[Pr(\|P_T\mathcal{R}_{\Omega}\mathcal{W}-\mathcal{W}\|_\text{max} > \frac{1}{2}\|\mathcal{W}\|_\text{max})\le  Pr(\langle P_T\mathcal{R}_{\Omega}\mathcal{W}-\mathcal{W},\mathcal{E}^{(v,i,j)}\rangle >\frac{1}{2}\|\mathcal{W}\|_\text{max} \text{ for any (v,i,j)} )N^2\]
For each $(v,i,j)$, sample ${s^\prime}={(v_{s^\prime},i_{s^\prime},j_{s^\prime})}$ according to the sampling distribution in Assumption $4$. 
Define $\Psi_{(v,i,j)}=\langle \mathcal{E}^{(v,i,j)}, P_T\mathcal{R}_{s^\prime}\mathcal{W} -\frac{1}{|\Omega|}\mathcal{W}\rangle$. 
Recall the definition of $\mathcal{R}_s$ from the paper. Now each entry of $P_T\mathcal{R}_{\Omega}\mathcal{W}-\mathcal{W}$ is distributed as $\sum_{s=1}^{|\Omega|} \Psi_{(v,i,j)}^{(s)}$, where $\Psi_{(v,i,j)}^{(s)}$ are iid samples of $\Psi_{(v,i,j)}$. 

We have that :
$|\Psi_{(v,i,j)}|\le \frac{1}{p(v,i,j)} \|P_T( \mathcal{E}^{(v,i,j)})\|_F^2\langle \mathcal{E}^{(v,i,j)},\mathcal{W}\rangle|\le \frac{1}{c^\prime\beta\log{N}}\|\mathcal{W}\|_\text{max}$

Also, $E[\Psi_{(v,i,j)}^2]= E[\frac{1}{p(v,i,j)^2}\langle \mathcal{E}^{(v,i,j)},\mathcal{W}\rangle^2\langle \mathcal{E}^{(v,i,j)}, \mathcal{E}^{(s^\prime)}\rangle^2]\le \frac{1}{|\Omega|c^\prime\beta\log{N}}$, where the expectation is over $s^\prime$. 
Standard Bernstein inequality can be used with the above bounds to prove the lemma.

\end{document}